\documentclass{article}

\usepackage[final]{neurips_2021_ml4ps}

\usepackage[utf8]{inputenc} 
\usepackage[T1]{fontenc}    
\usepackage{url}            
\usepackage{booktabs}       
\usepackage{amsfonts}       
\usepackage{nicefrac}       
\usepackage{microtype}      
\usepackage{xcolor}         
 \usepackage[pdftex]{graphicx} 
\usepackage{amsmath}
\DeclareMathOperator*{\argminA}{arg\,min} %
\usepackage{subcaption}
\usepackage{float}
\usepackage{hyperref}

\title{Discovering PDEs from Multiple Experiments}

\author{
  Georges Tod$^1$, Gert-Jan Both$^1$, Remy Kusters$^{1,2}$ \\
  $^1$Université de Paris, INSERM U1284, CRI, 75004 Paris, France \\
  $^2$IBM Research Paris-Saclay, 91400 Orsay, France\\
  \texttt{\{firstname.lastname\}@cri-paris.org}    }

\begin{document}

\maketitle

\begin{abstract}
Automated model discovery of partial differential equations (PDEs)  usually considers a single experiment or dataset to infer the underlying governing equations. In practice, experiments  have inherent natural variability in parameters, initial and boundary conditions that cannot be simply averaged out. We introduce a randomised adaptive group Lasso sparsity estimator to promote grouped sparsity and implement it in a deep learning based PDE discovery framework\footnote{\label{footnote:code} Data, code and results shared on: \url{https://github.com/georgestod/multi_deepmod}.}. It allows to create a learning bias that implies the a priori assumption that all experiments can be explained by the same underlying PDE terms with potentially different coefficients. Our experimental results show more generalizable PDEs can be found from multiple highly noisy datasets, by this grouped sparsity promotion rather than simply performing independent model discoveries.
\end{abstract}

\section{Introduction}
A classical approach to perform model discovery is by sparse regression and consists in finding $\xi$ such that, $u_t = \Theta \cdot \xi$, where $u_t$ is the time derivative of the field $u$. Each column of $\Theta$ is a candidate term of the underlying PDE, typically a combination of polynomial and spatial derivative functions (e.g. $u$, $u_x$, $uu_x$). Usually, $\xi$ is identified based on a single experiment consisting of $n$ samples of the field $u$, see \cite{brunton2016discovering,rudy2017,schaeffer2017,maddu2019stability,both2019,tod2021sparsistent}. However, in practice observations of the same experiment might lead to some measurement differences. For instance, it might not be possible: (1) to fix identically the initial and/or boundary conditions of several experiments, (2) some of the parameters might not be controllable and will vary. As a result, there will be experimental results which exhibit natural variability. Let us denote $q$ the number of experiments at hand from which we would like to find the underlying partial differential equations. If we perform $q$ individual model discoveries as described earlier, the data across experiments will not be leveraged, loosing an opportunity for learning. To leverage the data across experiments, the $q$ libraries could be stacked $\Theta = [\Theta_{1} , \Theta_{2}, \dots, \Theta_{q}]^T$ and then a single sparse regression could reveal $\xi$, see \cite{rudy2017,chen2020}. However, this approach is not adapted if the coefficients of the underlying PDEs vary from an experiment to another. 
In this paper, a more general approach to this is proposed: by promoting sparsity group-wise we leverage the data across experiments and are able to handle varying coefficients across datasets. In \cite{rudy2019parametric}, a group sequentially thresholded ridge regression was proposed to infer parametric PDEs from a single dataset. In \cite{maddu2020learning} a group Iterative Hard Thresholding algorithm is used to enforce conservation laws and impose symmetries. In \cite{de2019discovery} a group Lasso is used it to try to infer the law of gravitation from experimental ball drops.

Furthermore, experimental data might contain high noise levels and it is known that pure sparse regression based model discovery cannot handle noise levels above 5 $\%$, see \cite{rudy2017,maddu2019stability}. For such cases, deep learning model discovery frameworks, that combine deep neural networks, automatic differentiation with sparse regression excel, see \cite{both2019,chen2020,tod2021sparsistent}. \\

\paragraph{Contributions} (1) we introduce a randomised adaptive group Lasso to promote grouped sparsity to discover the common underlying PDE from multiple experiments. The group adaptive Lasso \cite{wei2010adaptiveGL} is known to have better asymptotic properties than the group Lasso \cite{yuan2006model} in terms of model selection consistency. The additional randomisation we propose leads empirically to better results, especially once integrated in a stability selection loop - (2) we implement the latter into the deep learning discovery framework DeepMod (see note \ref{footnote:code}) to perform discoveries from highly noisy datasets. Overall, our experimental results show for the first time, more generalizable PDEs can be found from multiple highly noisy datasets, by promoting grouped sparsity rather than simply performing independent model discoveries. Finally, our results suggest sharing layers across neural networks within DeepMod does not necessarily improve the inference performance.
\section{Methods}
\paragraph{Sparse regression based model discovery} An approach to discover PDEs from $q$ experiments is to repeat $q$ times a method used to perform discoveries from single experiments. Following \cite{tod2021sparsistent} a randomised adaptive Lasso to select the terms of the PDE can be used for $i^{\text{th}}$ experiment,
\begin{equation}
        \hat{\tilde{\xi}}_i = \argminA_{\tilde{\xi}} \Big( \frac{1}{2n}|| \partial_t u_i - \tilde{\Theta}_i \tilde{\xi}_i ||_{2}^{2} + \lambda \sum_{j=1}^{p}  \frac{||\tilde{\xi}_{i,j} ||_{1}}{W_{i,j}}  \Big)
        \label{eq:RadaLasso}
\end{equation}
where $\tilde{\Theta}_{i,j} = \Theta_{i,j}/\hat{w}_{i,j} $, $\tilde{\xi}_{i,j} = \hat{w}_{i,j} \cdot \xi_{i,j}$, $\hat{w}_{i,j} = 1/|\hat{\xi}_{i,j}|^\gamma$, with $\gamma=2$ and $W_{i,j}$ is randomly selected from a beta distribution, $w \sim \beta(1,2)$ to promote weights close to 0. While this randomised version\footnote{\label{footnote:randomisation} A lose connection can be made here with the transformation from decision trees to random forests: the randomisation will select features randomly.} of the adaptive Lasso is less common than its deterministic counter part, it is known to perform better in terms of variable selection consistency  \cite{meinshausen2010stability,tod2021sparsistent}. Similar observationsas in \cite{tod2021sparsistent} appear with grouped variables, see Appendix \ref{appendix:WHYadaptive}.

\paragraph{Grouped sparsity} By assuming the underlying PDE terms are the same across multiple experiments, with potentially different coefficients, grouped sparsity allows to leverage the data across the experiments. The previous estimator can be generalised (inspired by \cite{yuan2006model}) to grouped variables,
\begin{equation}
              \hat{\tilde{\xi}}= \argminA \Big( \frac{1}{2n}||\partial_t u - \tilde{\Theta} \tilde{\xi}||_{F}^{2} + \lambda \sum_{g=1}^{p} \frac{||\tilde{\xi}_g||_{2}}{W_{g}}  \Big)
\end{equation}
where $\tilde{\Theta}_{i,g} = \Theta_{i,g}/\hat{w}_{i,g}$, $\tilde{\xi}_{i,g} = \hat{w}_{i,g}\cdot \xi_{i,g}$, $\hat{w}_{i,g} = 1/|\hat{\xi}_{i,g}|^\gamma$ and $W_{g}$ is randomly selected from $w$. If there is one element per group, the randomised adaptive group Lasso\footnote{Implementation wise, the group Lasso python library MuTaR is used, see \cite{mutar}.} becomes the randomised adaptive Lasso. The amount of regularisation parametrized by $\lambda$ is found automatically using stability selection with some error control, see Appendix \ref{appendix:ss}.
\paragraph{Deep learning based model discovery}
In practice, experimental data contains not only natural variability but also noise. It is known that pure sparse regression based model discovery cannot handle high noise levels (>5 $\%$) - for such cases, deep learning model discovery frameworks excel see \cite{both2019,chen2020} . Here, we extend DeepMod \cite{both2019} to handle multiple experiments (see note \ref{footnote:code}). The framework combines a function approximator of $u$, typically a deep neural network which is trained with the following loss, 
\begin{equation}
\mathcal{L} = \sum_{i=1}^{q} \Big(  \underbrace{\frac{1}{n} ||u_i-\hat{u}_i ||_{2}^{2}}_{\mathcal{L}_{\textit{MSE}}} + \underbrace{\frac{1}{n} ||\partial_t \hat{u}_i - \Theta_i (\hat{\xi}_i \cdot M_i) ||_{2}^{2}}_{\mathcal{L}_{\textit{reg}}}   \Big)
\label{eq:deepmod}
\end{equation}
The first term $\mathcal{L}_{\textit{MSE}}$ learns the data mapping $(x, t) \to \hat{u}_i$, while the second term $\mathcal{L}_{\textit{reg}}$ constrains the function approximator to solutions of the partial differential equation given by $\partial_t u_i, \Theta_i$ and $(\hat{\xi}_i\cdot M_i)$. The terms to be selected in the PDEs are determined using a mask $M$ derived from the result of the randomised adaptive Lasso/group Lasso,
\begin{equation}
    M_{i} = \left\{
    \begin{array}{ll}
        1     & \text{if } \tilde{\xi}_j \in S_{\text{stable}}^{\Lambda^*} \\
	0     & \text{otherwise}
    \end{array}
\right.
\label{eq:mask}
\end{equation}
where $j$ is the index of a potential term and $S_{\text{stable}}^{\Lambda^*}$ is determined by equation (\ref{eq:stable_set}). The coefficients $\hat{\xi}$ in front of the potential terms are computed using a Ridge regression on the masked library $(\Theta_i \cdot M_i)$. During training, if $\sum_{i=1}^{q} \mathcal{L}_{\textit{MSE}}$ on the test set does not vary anymore or if it increases, the sparsity estimator is triggered periodically. As a result, the PDE terms are selected iteratively by the dynamic udpate of the mask $M$ during the training.
%
%
%
\section{Experiments} 
\begin{figure}[H]
    \centering
     \begin{subfigure}[b]{0.45\textwidth}
     	\centering
    	\includegraphics[width=6cm]{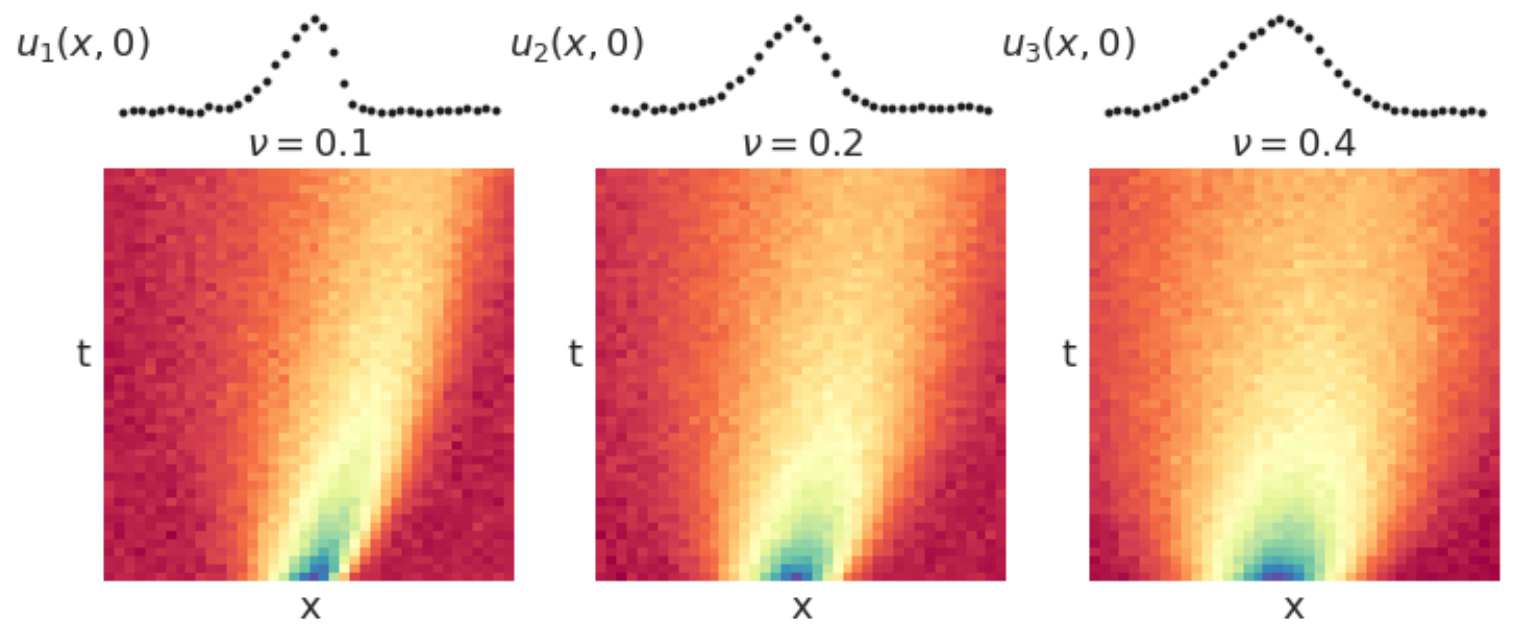}
    	\caption{\textit{case 1: with different parameters $\nu$.}}
     \end{subfigure}    
     \begin{subfigure}[b]{0.45\textwidth}
     	\centering
    	\includegraphics[width=6cm]{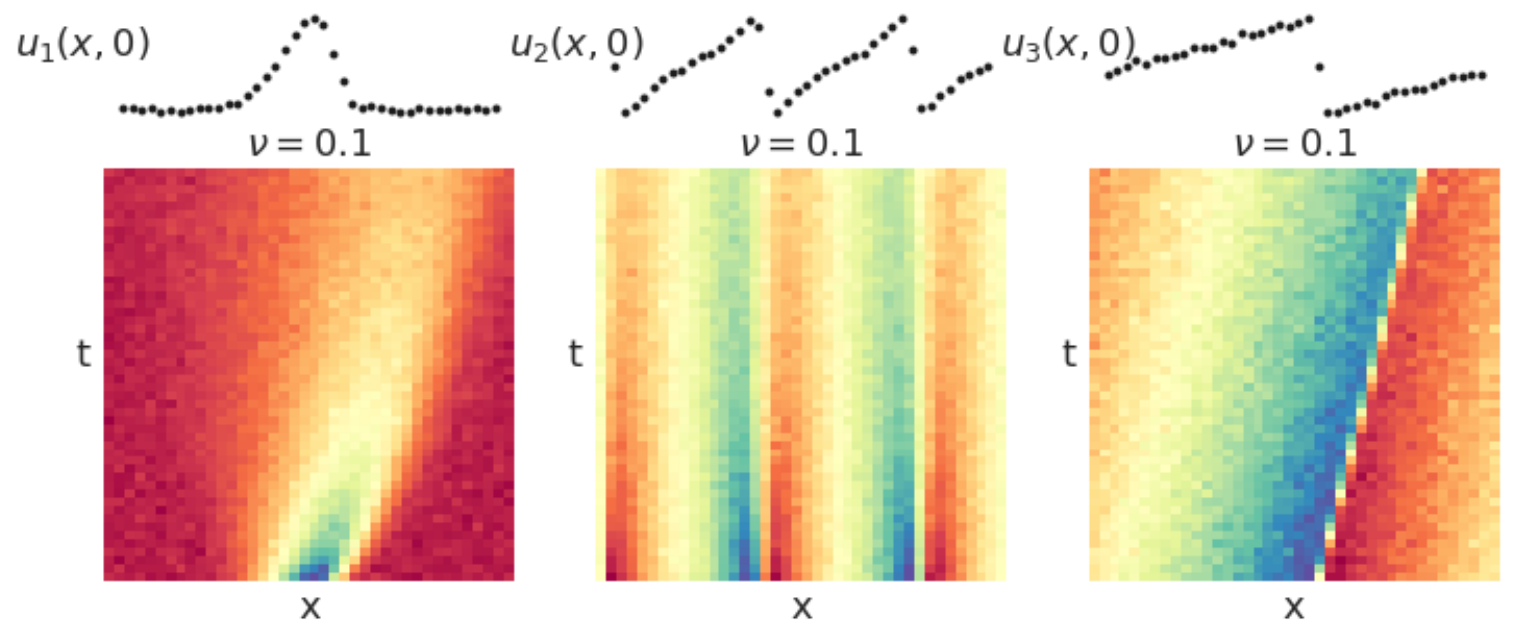}
    	\caption{\textit{case 2: with different initial conditions.}}
     \end{subfigure}    
    \caption{\textit{Cases 1 and 2 datasets.} Burgers' PDE is solved analytically for (a) different values of the viscosity $\nu$ and (b) the initial conditions vary at fixed viscosity. Each case dataset contains 3 $\times$ 2000 samples with the addition of $10 \%$ Gaussian white noise.}
    \label{fig:datasets}
\end{figure}

\begin{figure}[H]
    \centering
    \includegraphics[width=14cm]{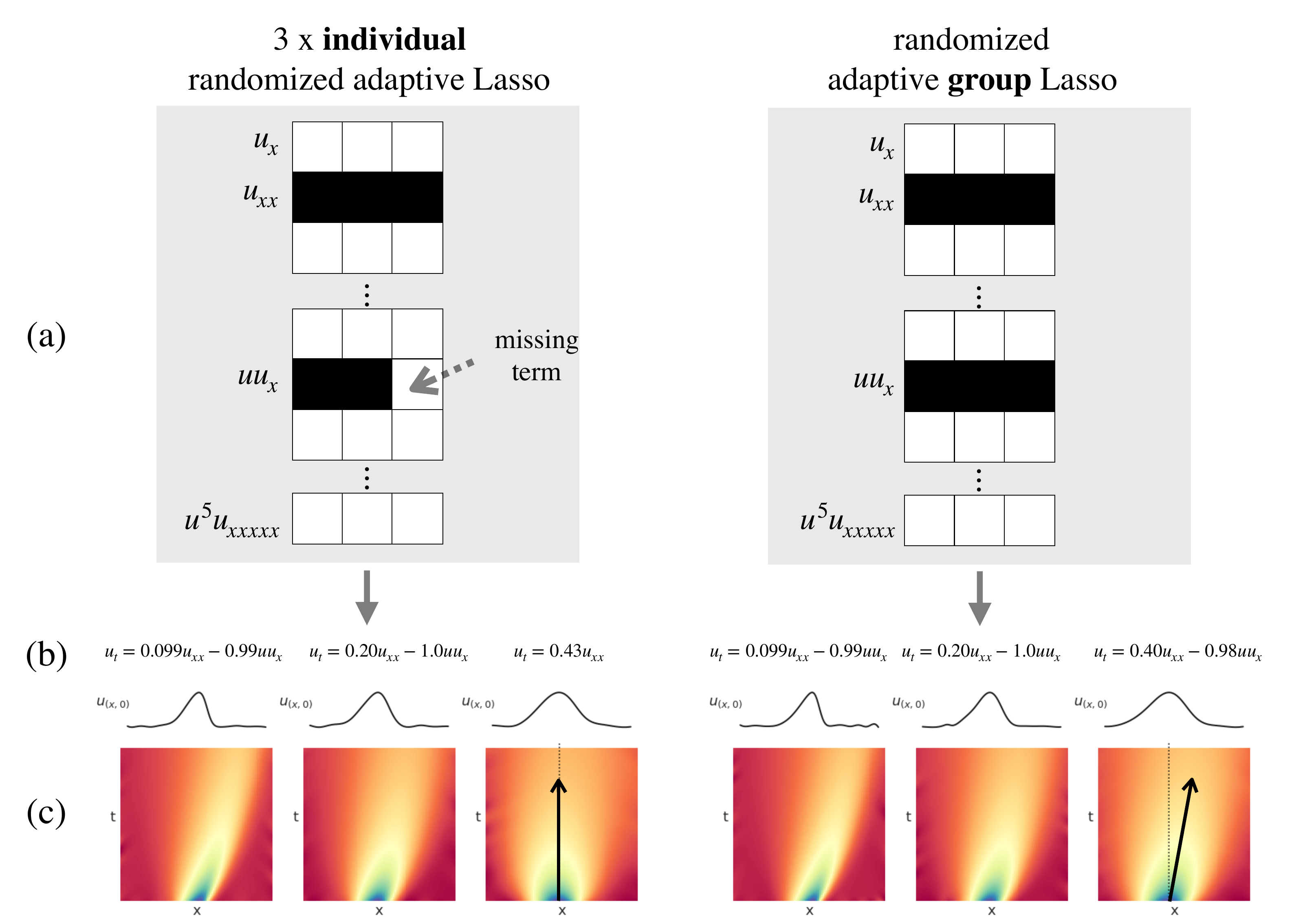}
    \caption{\textit{Example of sparsity patterns (a), equations recovered (b) and data interpolations (c) by DeepMod from case 1 dataset}. In (b) the absence of the nonlinear advection term ($uu_x$) results on the recovery of a PDE with pure diffusive trajectory (c). Using the randomised adaptive group Lasso allows to enforce a learning bias that helps finding the actual ground truth: the Burgers' equation ($u_t= \nu u_{xx} -uu_x$) with varying viscosities $\nu$.}
    \label{fig:params}
\end{figure}

\begin{figure}
    \centering
     	\includegraphics[height=5cm]{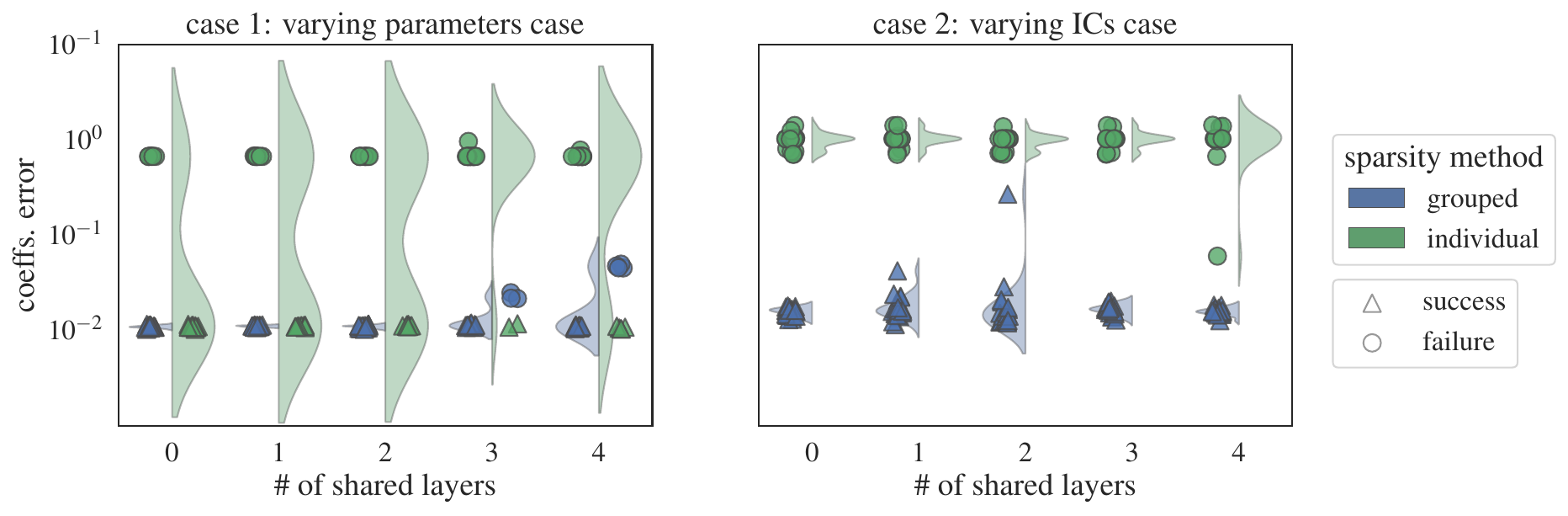}
    \caption{\textit{Performance of multitask task learning in model discovery}. Each experiment is repeated with 20 different initialisation seeds for each NN architecture. By promoting sparsity as a group and without sharing layers across neural networks, DeepMod becomes a sharp estimator of the common underlying PDE from multiple experiments.}
    \label{fig:shared_layers}
\end{figure}

In this section, we first present two cases based on the viscous Burgers' equation: $u_t= \nu u_{xx} -uu_x$. The first case consists in generating three spatiotemporal datasets with an identical initial condition (Dirac delta) and different viscosity parameters $\nu = \{0.1,0.2,0.4\}$, see figure \ref{fig:datasets}(a). For the second case we fix $\nu=1$ and vary the initial conditions (Dirac delta, periodic and step-like functions) see figure \ref{fig:datasets}(b). These datasets are generated from analytical solutions \cite{tod2021sparsistent}, on top of which we add a high noise level of $10 \%$ Gaussian white noise. In both cases, 40 points in time and 50 points in time are considered leading to 2000 samples per experiment. In such conditions, the observations of the fields $u$ are coarsely sampled and noisy enough such that it would not be possible to compute accurate enough derivatives for the libraries using classical methods such as finite differences or polynomial interpolation. By leveraging automatic differentiation and physics informed neural networks to compute the libraries, these cases are typically the ones where deep learning based model discovery frameworks excel.  We compare the two different methods for promoting sparsity in DeepMod described in the previous sections: the individual sparsity promotion method is performed $q$ times (once per experiment) or the grouped sparsity promotion is performed, see figures \ref{fig:params} and \ref{fig:shared_layers}. The sparsity patterns that are recovered might be different, see figure \ref{fig:params}(a): the individual randomised adaptive Lasso is incapable of recovering the nonlinear advection term for the last experiment. This can be explained physically, as it happens for the experiment with the highest diffusion coefficient ($\nu=0.4$): diffusion is predominant over nonlinear advection. On the other hand, by promoting grouped sparsity, the ground truth underlying PDE can be recovered and it can be verified that it actually generalises better since $\text{MSE}_{\text{test,group}} = 1.1749e^{-3} < \text{MSE}_{\text{test,individual}} = 1.1761e^{-3}$. We report a similar conclusion for the chaotic Kuramoto-Sivashinsky equation, $\text{MSE}_{\text{test,group}} = 5.51e^{-3} < \text{MSE}_{\text{test,individual}} = 5.59e^{-3}$, see figure \ref{fig:KS}. The same hyperparameters were used, see Appendix \ref{appendix:hyperparams} and 2 $\times$ 2000 points are randomly sampled from a numerical solution \cite{rudy2017}, on top of which we add a high noise level of $20 \%$ Gaussian white noise. Since we use NNs to interpolate across samples, we also investigated the impact of sharing NN layers across tasks for cases 1 and 2. Interestingly, sharing layers across neural networks does not help DeepMod finding better coefficient estimates nor improves the variable selection performance - it can actually be the opposite, see figure \ref{fig:shared_layers} and Appendix \ref{appendix:multi_NNs}.

It is important to keep in mind when promoting grouped sparsity, that if a term is important to recover a PDE, it will be forced to be conserved for all the other PDEs. By doing so, we create a learning bias that implies the a priori assumption that all experiments can be explained by the same underlying PDE terms with potentially different coefficients. Obviously, the learning bias can also turn into a \textit{negative learning transfer} if the latter assumption is wrong. A limitation of our approach is reported when the discovery problem does not have unique solutions, see Appendix \ref{appendix:more_cases}.
\begin{figure}[H]
    \centering
     \begin{subfigure}[b]{0.48\textwidth}
     	\centering
    	\includegraphics[height=3.3cm]{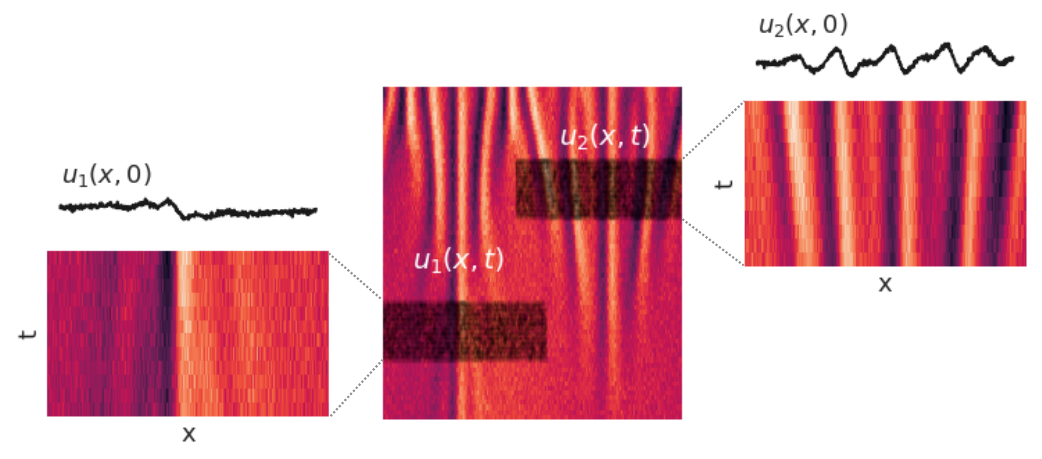}
    	\caption{}
     \end{subfigure}    
     \begin{subfigure}[b]{0.48\textwidth}
     	\centering
    	\includegraphics[width=6cm]{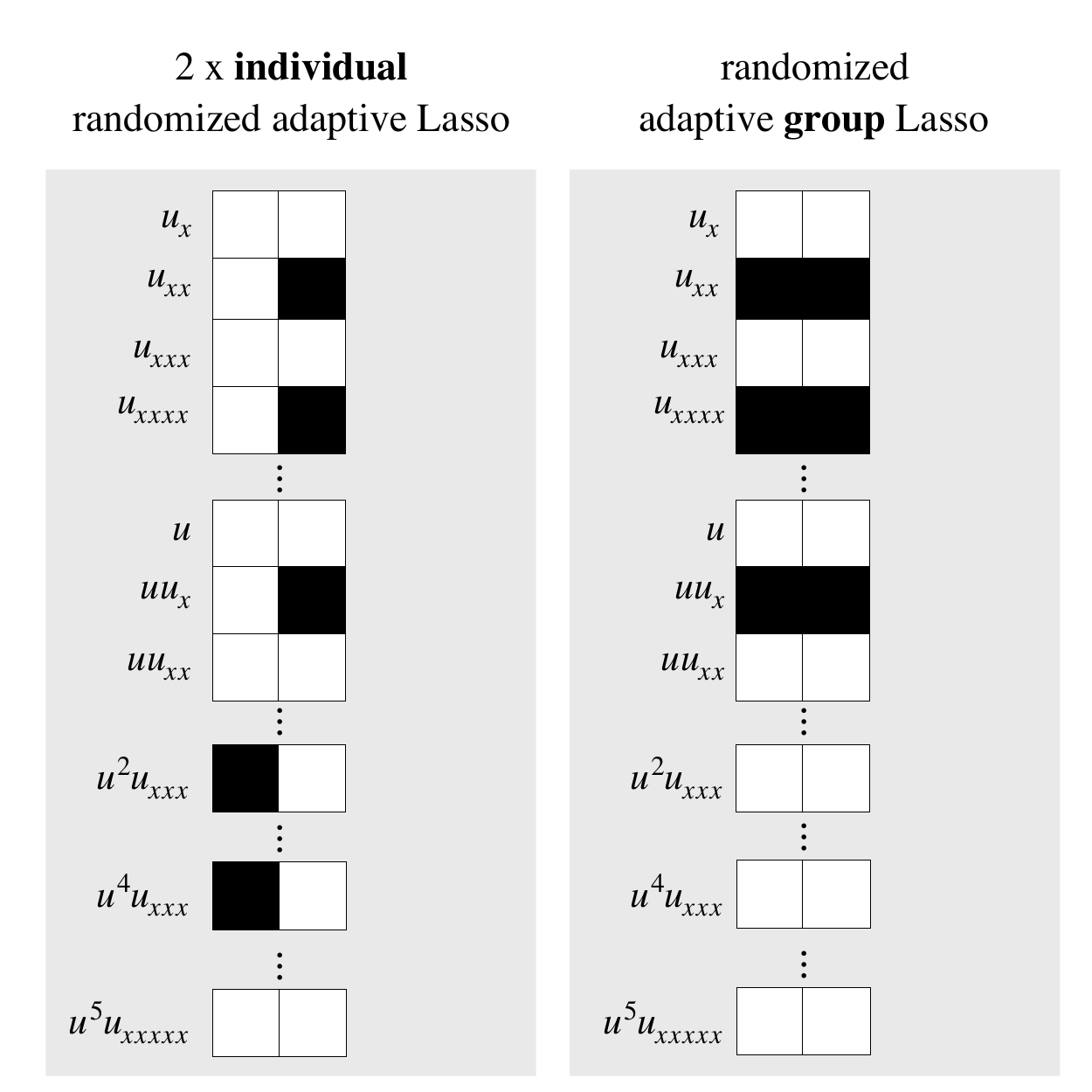}
    	\caption{}
     \end{subfigure} 
    \caption{\textit{Case 3 dataset and sparsity patterns recovered by DeepMod.} In this experiment, we extract two datasets (2 $\times$ 2000 random samples) in (a) pre-chaotic ($u_1$) and chaotic ($u_2$) regimes from a numerical solution \protect\cite{rudy2017} of the Kuramoto-Sivashinsky equation ($u_t =  -uu_x -u_{xx} - u_{xxxx}$), with $20\%$ Gaussian white noise. (b) The underlying PDE can only be recovered when grouped sparsity is promoted.}
    \label{fig:KS}
\end{figure}
\section{Conclusion}
We extend the limitation of relying on a single experiment to infer PDEs and explore how to leverage the information across multiple experiments with different parameters, initial and/or boundary conditions. To do so, we first introduce a randomised adaptive group Lasso to promote grouped sparsity. This creates a learning bias in the model discovery process which assumes there is a common underlying PDE with potentially different coefficients across experiments. Second, once integrated in the deep learning model discovery framework DeepMod, our results show \textit{more generalizable PDEs} can be found from highly noisy datasets, by promoting grouped sparsity rather than simply performing independent model discoveries.

Future work will focus on the extension of the approach to additional spatial dimensions in order to tackle real life experimental applications such as retrieving recovering fluid flow equations in turbulent conditions from particle image velocimetry.
\begin{ack}
This work received support from the CRI Research Fellowship to attributed to Remy Kusters.
\end{ack}

\bibliographystyle{unsrt}  
\typeout{}
\bibliography{main}

\appendix
\section{Group Lasso and Adaptive Group Lasso}
\begin{figure}
    \centering
     \begin{subfigure}[b]{0.48\textwidth}
     	\centering
    	\includegraphics[height=4cm]{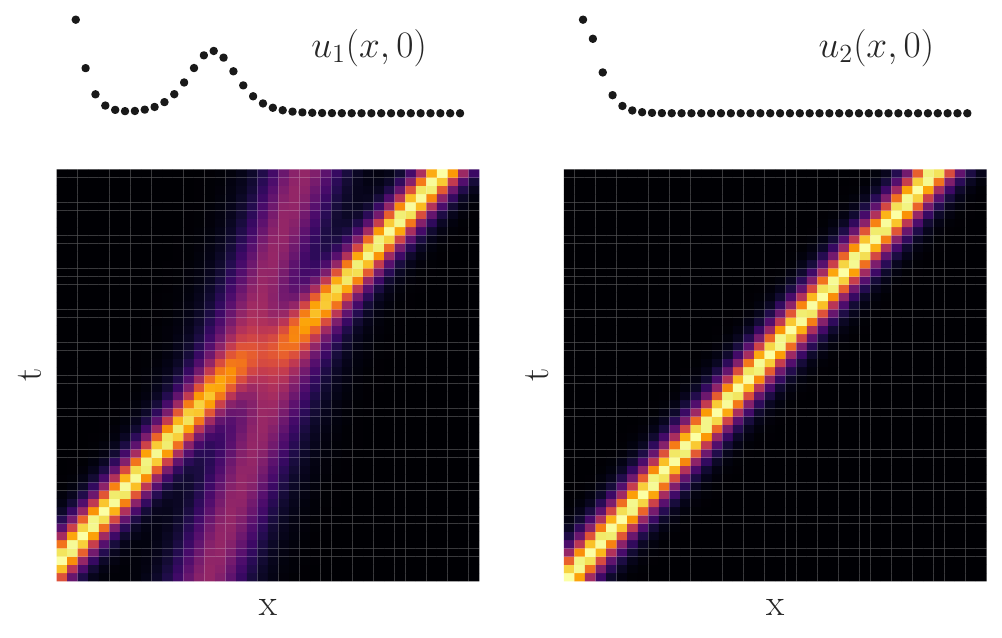}
    	\caption{}
     \end{subfigure}    
     \begin{subfigure}[b]{0.48\textwidth}
     	\centering
    	\includegraphics[width=6.5cm]{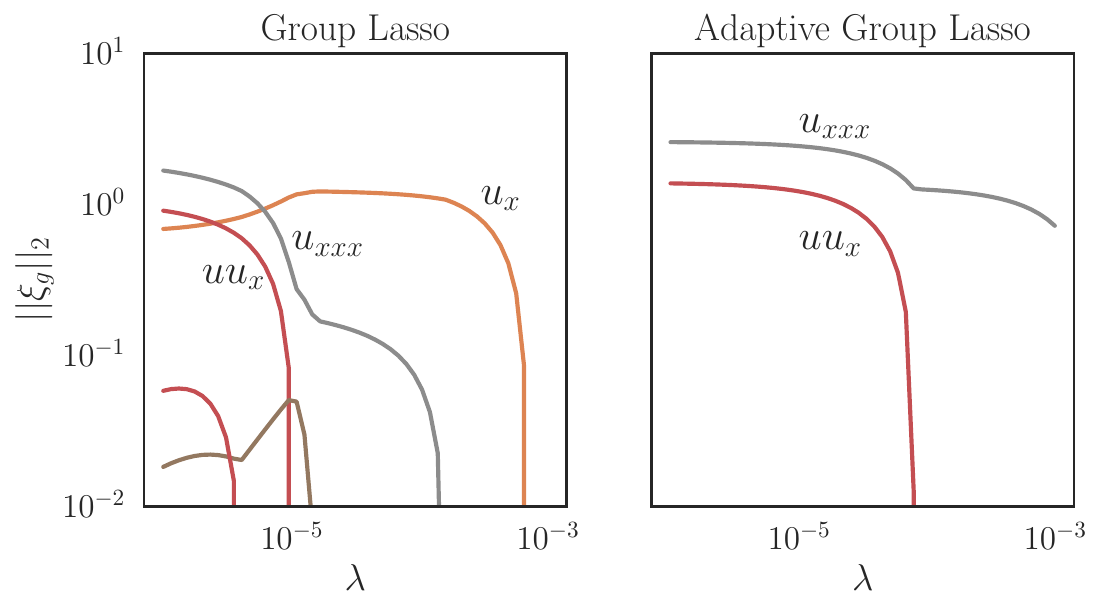}
    	\caption{}
     \end{subfigure}    
    \caption{\textit{When the group Lasso fails to recover the Kortweg-de-Vries equation ($u_t = -6 uu_x - u_{xxx}$)}. From $2 \times 36$ terms libraries computed from two analytical solutions with single and double solitons. Group norms of coefficients paths show the group Lasso would not manage to find the KdV equation for any $\lambda$, while the Adaptive Group Lasso would.}
    \label{fig:Kdv_noiseless}
\end{figure}

An illustration of when the group Lasso might not recover the underlying PDEs where the adaptive group Lasso would is given, see see figure \ref{fig:Kdv_noiseless}. We find the irrelevant term $u_x$ is highly correlated with the relevant term $u_{xxx}$. In that situation, the group Lasso would not manage to find the KdV equation for any $\lambda$, see figure \ref{fig:Kdv_noiseless}, while the Adaptive Group Lasso would. An idea of why such a behavior can be found in \cite{tod2021sparsistent}.

\label{appendix:WHYadaptive}

\section{Stability selection for the randomised adaptive Lasso/group Lasso}
\label{appendix:ss}
This appendix extends stability selection with error control from \cite{tod2021sparsistent} to the randomised adaptive group Lasso.
With stability selection, variables are chosen according to their probabilities of being selected with some warranty on the selection error, \cite{meinshausen2010stability}. The first step consists in finding the probability of a variable $k$ of being selected under some data perturbation: let $I_{b}$ be one of $B$ random sub-samples of half the size of the training data drawn without replacement. For a given $\lambda$ an estimation of the probability of $k$ being selected is given by, 
\begin{equation}
    \hat{\Pi}^{\lambda}_{k} = \frac{1}{B}\sum_{b=1}^{B} \left\{
    \begin{array}{ll}
        1     & \text{if } f_{k}^{\lambda}(I_b) > 0 \\
		0 		& \text{otherwise}
    \end{array}
\right.
\end{equation}
where $f_{k}^{\lambda}(I_b) = ||\hat{\xi}^{\lambda}_{k}({I_b})||_1$ for the randomised adaptive Lasso. The extension to the  randomised adaptive group Lasso simply requires $f_{k}^{\lambda}(I_b) = ||\hat{\xi}^{\lambda}_{k}({I_b})||_2$. Second, by computing the probabilities of being selected over a given range of $\lambda$'s the stability paths can be obtained. That range is initially denoted $\Lambda = [\epsilon \lambda_{max}; \lambda_{max}]$, where $\epsilon$ is the path length and  $\lambda_{max}$ is the regularisation parameter where all coefficients $\hat{\xi}$ are null. \cite{meinshausen2010stability} derive an upper bound on the expected number of false positives $\mathbb{E}(V)$, that can help determining a smaller $\Lambda$ region where some control on the selection error can be warrantied. By fixing this bound to $EV_{max}$, the regularisation region becomes,
\begin{equation}
    \Lambda^* = \left\{\lambda \in \Lambda \text{ such that, } \mathbb{E}(V) \leq \frac{q_\Lambda^2}{(2\pi_{thr}-1)p} \leq EV_{max} \right\}
    \label{eq:BigLambda}
\end{equation}
where $\pi_{thr}$ is the minimum probability threshold to be selected and $q_\Lambda$ is the average of selected variables that we propose to approximate by $\hat{q}_\Lambda = \sum_k \hat{\Pi}^{\lambda}_{k}$. Finally, the set of stable variables with an upper bound on the expected number of false positives is,
\begin{equation}
    S_{\text{stable}}^{\Lambda^*} = \left\{k \text{ such that } \max \hat{\Pi}^{\lambda}_{k} \geq \pi_{thr} \text{ for } \lambda \in \Lambda^* \right\} 
        \label{eq:stable_set}
\end{equation}
\section{A single set of hyperparameters}
\label{appendix:hyperparams}
A single set of hyperparameters is used for all cases presented in this work - except for the experiments presented in figure \ref{fig:shared_layers} where the NN architectures vary. 
\paragraph{Stability selection}  expected number of false positives upper bound $EV_{max}=3$, number of resamples $B=40$ and the minimum probability to be selected $\pi_{thr}=0.9$.

\paragraph{Library} consists of polynomials and partial derivatives up to the fifth order leading to a library size of $p=$36 potential terms.
\paragraph{Neural network architecture \& optimiser} NNs are 4 layers deep with 65 neurons per layer and sinus activation functions with a specific initialisation strategy, see \cite{sitzmann2020implicit}. The NNs are trained by an Adam optimiser with a learning rate of $5 \cdot 10^{-5}$ and $\beta = (0.99,0.99)$.
\section{Multitask neural networks performance}
\label{appendix:multi_NNs}
When sharing layers, the samples used to train for a task are going to be used in the training of other tasks as well, creating another learning bias. To verify more systematically the performance of the promotion of grouped sparsity and multitask neural networks, we vary the methods to promote sparsity and the NN architectures \footnote{ To make sure the NN architectures are comparable, the total number of weights and biases $\sim 40k$ are fixed.}, see figure \ref{fig:shared_layers}. 

\paragraph{Performance metrics} We introduce two metrics: (1) the result of an experiment is declared as a success, if and only if, the terms of the ground truth underlying PDE are found. In the case of failure, the metric is uninformative about whether an additional term was discovered or a term was missing - (2) the coefficients error metric quantifies the discrepancy between the right hand side coefficients of the ground truth PDEs and the discovered ones by computing: $||\xi - \hat{\xi}||_{F}/||\xi||_{F}$, where $F$ stands for the Frobenius norm.

Overall the variance of the results is much lower when promoting sparsity as a group. When no layers are shared, the coefficients error is the lowest and the success rate of the discoveries is equal to 1 when the grouped sparsity method is leveraged. Furthermore, the common underlying PDE across experiments with varying ICs can only be found when grouped sparsity is promoted. Interestingly, sharing layers across neural networks does not help DeepMod finding better coefficient estimates nor improves the variable selection performance - it can actually be the opposite. 
\section{Non-unique solutions limitation}
\label{appendix:more_cases}
\begin{figure}
    \centering
     \begin{subfigure}[b]{0.48\textwidth}
     	\centering
    	\includegraphics[height=4cm]{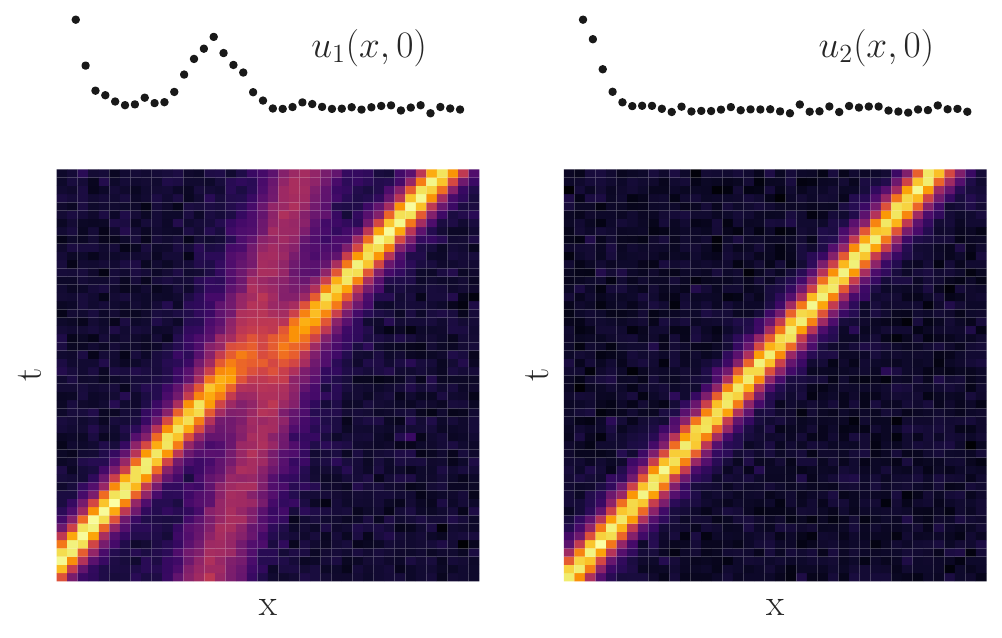}
    	\caption{}
     \end{subfigure}    
     \begin{subfigure}[b]{0.48\textwidth}
     	\centering
    	\includegraphics[width=6.5cm]{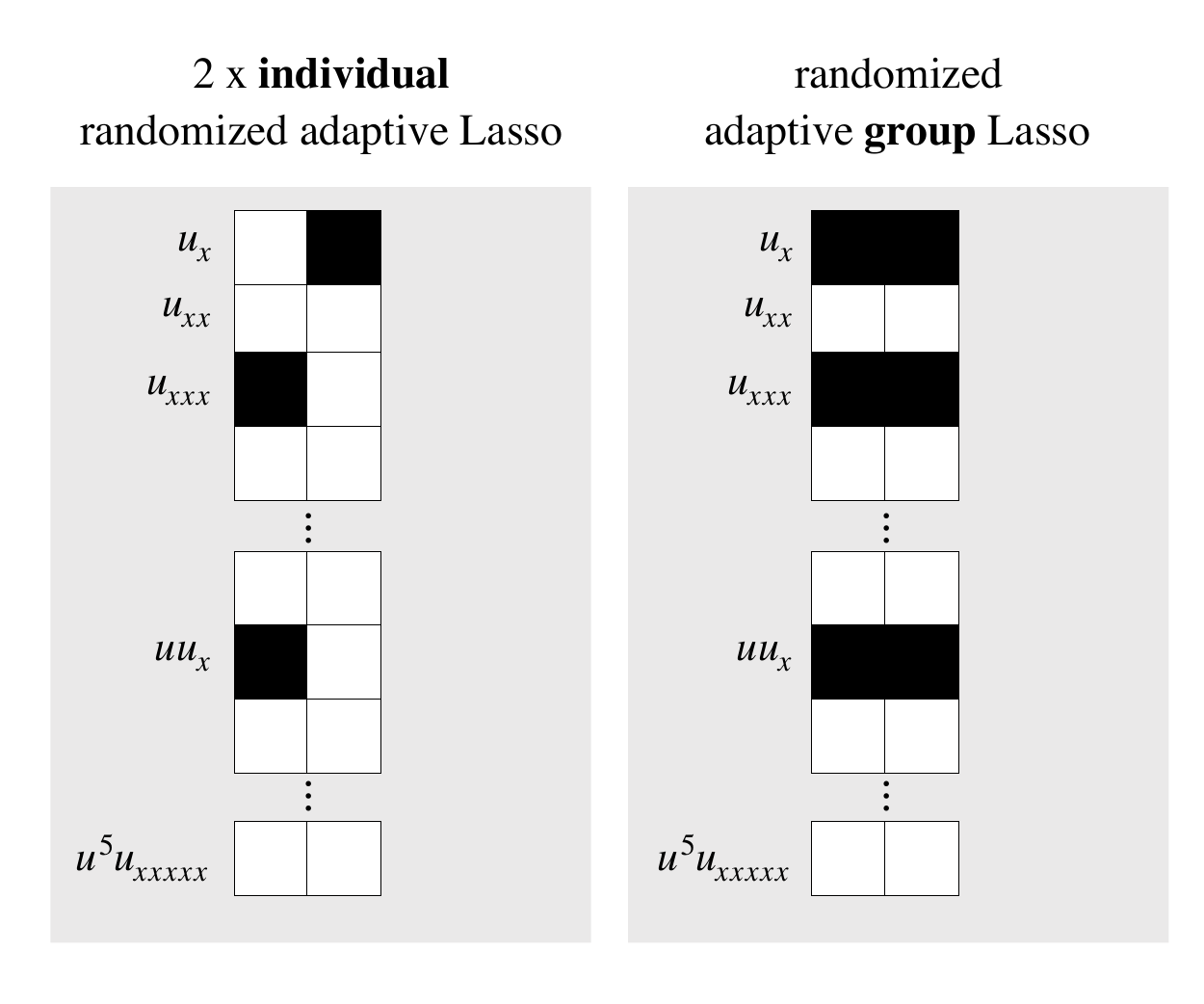}
    	\caption{}
     \end{subfigure}    
    \caption{\textit{Case 4 dataset and sparsity patterns recovered by DeepMod.}. In this experiment, we infer the equations from two analytical solutions of the Kortweg-de-Vries equation ($u_t = -6 uu_x - u_{xxx}$): with single and double solitons. The dataset consists of 2 $\times$ 2000 samples with the addition of $10 \%$ Gaussian white noise.}
    \label{fig:Kdv}
\end{figure}

In this experiment, we infer the equations from two analytical solutions of the Kortweg-de-Vries equation ($u_t = -6 uu_x - u_{xxx}$): with single and double solitons, see figure \ref{fig:Kdv}(a). Each dataset consists of 2 $\times$ 2000 samples with $10 \%$ Gaussian white noise. The resulting sparsity pattern, see figure \ref{fig:Kdv}(b), from the individual sparsity promotion is correct: the single soliton solution of the KdV equation is also a solution of the simpler travelling wave equation: $u_t = - c \cdot u_x$, where $c$ is the velocity of the wave. This illustrates the non-uniqueness of solutions of the PDE discovery problem. The promotion of grouped sparsity results in the discovery of a PDE with the 3 terms: $u_x,uu_x$ and $u_{xxx}$. While we know the single soliton is a solution of a PDE containing those 3 terms, the double soliton solution is not. The spurious term $u_x$ can be discerned by remarking its small magnitude: $u_t = -0.083 u_x - 5.9 uu_{x} - 0.99u_{xxx}$ - alternatively, it can also be diagnosed by comparing MSEs on some out of training data: $\text{MSE}_{\text{test,group}} = 5.48e^{-4}>\text{MSE}_{\text{test,individual}} = 5.41e^{-4}$. However, without the addition of noise, the library $\Theta$ can be constructed analytically and the randomised adaptive group Lasso does find the KdV equation as the underlying PDE for both datasets. We speculate the origin of the limitation of DeepMod in the case of non-unique solutions, comes from the Ridge regression used to compute the regularisation term of the loss, see equation \ref{eq:deepmod}. As a matter of fact, we remark for this experiment only, the results from DeepMod are very sensitive to the regularisation parameter fixed for the Ridge regression.
\end{document}